\documentclass[twoside,11pt]{article}

\usepackage{blindtext}
\usepackage{jmlr2e}
\usepackage{graphicx} 
\usepackage{listings}
\usepackage{subcaption}
\usepackage{amsmath}
\usepackage{amsfonts}
\usepackage{amssymb}
\usepackage{color}
\usepackage{xcolor}
\usepackage{blindtext}
\usepackage{url}
\usepackage{enumitem}
\usepackage{cite}
\usepackage{comment}
\usepackage{ulem}
\usepackage{array}
\usepackage{booktabs}
\usepackage{multirow}

\setitemize{noitemsep,topsep=02pt,parsep=02pt,partopsep=02pt}

\usepackage{natbib}

\begin{document}

\title{PyPulse: A Python Library for Biosignal Imputation}

\author{\name Kevin Gao \email kevingao26@gmail.com\\
        \addr Independent Researcher
        \AND
        \name Maxwell A. Xu \email maxu@illinois.edu\\
        \addr University of Illinois Urbana-Champaign
        \AND
        \name James M. Rehg \email jrehg@illinois.edu\\
        \addr University of Illinois Urbana-Champaign
        \AND
        \name Alexander Moreno \email alexander.f.moreno@gmail.com\\
        \addr Independent Researcher
        }
\editor{}

\maketitle

\begin{abstract}
We introduce PyPulse, a Python package for imputation of biosignals in both clinical and wearable sensor settings. Missingness is commonplace in these settings and can arise from multiple causes, such as insecure sensor attachment or data transmission loss. PyPulse's framework provides a modular and extendable framework with high ease-of-use for a broad userbase, including non-machine-learning bioresearchers. Specifically, its new capabilities include using pre-trained imputation methods out-of-the-box on custom datasets, running the full workflow of training or testing a baseline method with a single line of code, and comparing baseline methods in an interactive visualization tool. We released PyPulse under the MIT License on Github and PyPI. The source code can be found at: \url{https://github.com/rehg-lab/pulseimpute}.
\end{abstract}


\section{Introduction}
Biosignal sensor technologies have revolutionized health monitoring and interventions, enabling real-time monitoring of vital physiological signals \citep{stuart2022wearable, abidoye2011using, da2018internet}. However, a primary challenge in utilizing signals for health monitoring is their susceptibility to missing data due to sensor displacement, technical faults, or interference \citep{rahman2017mdebugger}. Traditional imputation methods, such as mean filling or linear interpolation \citep{le2018mean, dong2019lininterp}, often fail to adequately address the unique characteristics of pulsative signals, such as their quasi-periodicity and the specific morphological features indicative of clinical significance. Additionally, the lack of large-scale, publicly available datasets with realistic missingness patterns has stalled the development and testing of advanced imputation methods.

PulseImpute \citep{xu2023pulseimpute} addresses this gap as the first work to provide realistic mHealth scenarios and a diverse set of baselines, including a novel transformer-based architecture that captures domain-specific information such as local context through dilated convolutions. However, while these efforts have provided invaluable insight, their work is not easily extendable by non-machine-learning researchers, despite the clear clinical applications. The codebase does not provide an API and has a highly specialized software stack, making it difficult to extend for user-provided custom datasets. For example, a given experiment configuration file is specific to a given dataset, missingness mechanism, and model, with no modularity, making it infeasible use a custom dataset or try experimental combinations of models and missingness mechanisms without significant structural changes to the codebase.

\begin{table}[t]
\centering
\resizebox{\textwidth}{!}{%
\footnotesize
\begin{tabular}{>{\raggedright\arraybackslash}p{0.16\textwidth}>{\raggedright\arraybackslash}p{0.24\textwidth}>{\raggedright\arraybackslash}p{0.74\textwidth}}
\toprule
\textbf{User Type} & \textbf{Focus Area} & \textbf{Use Cases} \\
\midrule
Health Researcher & Upload noisy signals to obtain cleaned dataset for downstream analysis. & PPG/ECG Imputation for HRV analysis \citep{benchekroun2023impact, tsai2021coherence};  Wearable signal imputation for enabling Just-in-Time-Interventions \citep{chatterjee2020smokingopp, ertin2011autosense,nahum2018just} \\
\midrule
Machine Learning Researcher & Research in general imputation methodology and theory.  & General temporal imputation methods \citep{cao2018brits}; Theoretical work on MAR, MCAR, MNAR mechanisms \citep{pedersen2017missing}; Uncertainty Modeling to propagate uncertainty to downstream classification \citep{jun2020uncertainty, hetvae} \\
\midrule
Interdisciplinary Researcher (Health and ML) & Develop novel imputation algorithms for health time-series applications. & FFT imputer and BDC transformer for pulsative signal imputation \citep{rahman2015fft, xu2023pulseimpute}; ECG-specific imputation \citep{varim,rpca,manimekalai2018knbp}; PPG quality assessment \citep{naeini2019real, orphanidou2018quality}\\
\bottomrule
\end{tabular}
}
\caption{Intended audience and their use cases for our PyPulse package}
\label{tab:audience}
\end{table}

In this paper, we introduce PyPulse, a Python package that allows the user to fully customize the process of imputation generation for missing data in a given dataset. PyPulse supports a broad range of scientific and engineering related use cases, as shown in Table \ref{tab:audience}. To support these uses, we have created a modularized software stack that provides new capabilities with an API framework: (i) a flexible and user-friendly configuration file stemming from structural changes to every module; (ii) support for custom datasets; (iii) support for custom missingness mechanisms and the ability to select any missingness mechanism with any dataset; (iv) a visualization module to compare the generated imputations of baseline methods on a single interactive plot. At a high level, PyPulse allows a user to run the full workflow by simply running a single command, while optionally uploading a custom dataset or modifying an experimental configuration file.

\section{Software Architecture}
\begin{figure}[t]
\includegraphics[width=\textwidth]{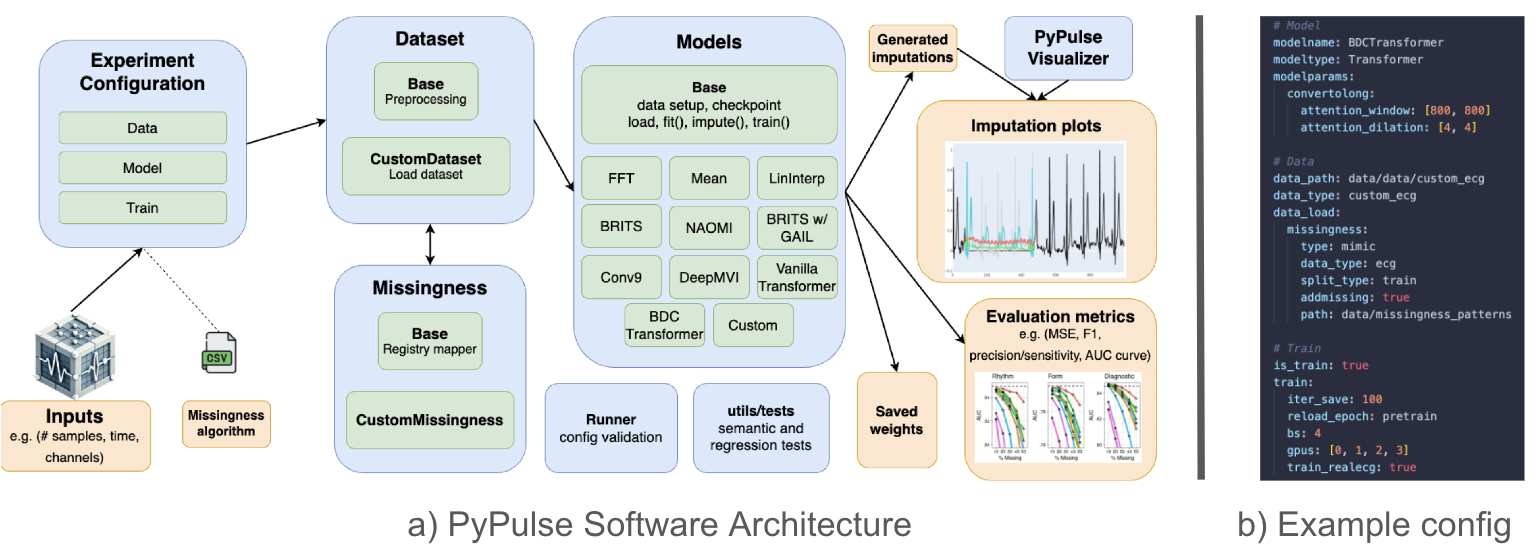}
\vspace{-.7cm}
\centering
\caption{a) PyPulse framework where blue are the modules, green are the classes, and orange is I/O. b) Example experiment config demonstrating model, data, and train inputs.}
\label{figure:arch}
\end{figure}
Figure~\ref{figure:arch}a shows the software architecture of PyPulse and each module is described below.

\vspace{0.75em}
\noindent\textbf{Experiment Configuration}\indent The \texttt{Configuration} module provides an organized hierarchy of YAML files sorted by model type and experiment name. Each configuration file is sorted into three sections: data, model, train. The \textit{data} section includes the the specific dataset being loaded along with its preprocessing as well as the missingness mechanism to be applied. The \textit{model} section specifies the imputation model and model-specific parameters. The \textit{train} section applies training hyperparameters (i.e. batch size). An example of this config can be found in Fig. \ref{figure:arch}b. The configuration files are validated before they are run, and they can be defined via command-line arguments or via a specific YAML file. 

\vspace{0.75em}
\noindent\textbf{Datasets}\indent In PyPulse, flexibility and extendability are achieved through a modular class hierarchy for managing datasets. The abstract base \texttt{Datasets} class manages the preprocessing of general datasets, while specialized subclasses can be defined for loading specific datasets with particular pre-processing necessities (e.g. denoising). The independent \texttt{Missingness} module defines a base class that maps missingness strategies specified in the configuration to their corresponding implementation classes. To facilitate customization, users can extend the \texttt{Datasets} or \texttt{Missingness} base class by overriding their respective preprocessing, data loading methods, and missingness application functions, allowing users to easily handle their unique datasets. Our custom dataset class supports various input formats out of the box. This class-based design ensures that new datasets and missingness methods can be seamlessly integrated without altering existing components.

\vspace{0.75em}
\noindent\textbf{Models}\indent \texttt{Models} provides a suite of 11 imputation algorithms that cover both classical and deep learning methodologies. Deep learning include the state-of-the-art BDC transformer method \citep{xu2023pulseimpute} and other transformer-based methods, DeepMVI \citep{deepmvi}, Vanilla \citep{originaltransformer}, Conv9 \citep{convattn}, RNN-based BRITS \citep{brits}, and RNN-GAN-based BRITS with GAIL \citep{gail} and NAOMI \citep{NAOMI}. Classical include the neighbor-based linear interpolation \citep{dong2019lininterp}, statistics-based mean filling \citep{le2018mean}, and the frequency-based FFT \citep{rahman2015fft}. The imputation, data loading, and training functionalities are handled generically in the base class for each category of model, and the addition of a specific novel model requires only the re-implementation of the forward pass.

\begin{figure}[!hbtp]
    \vspace{-.8cm}
    \includegraphics[width=\textwidth]{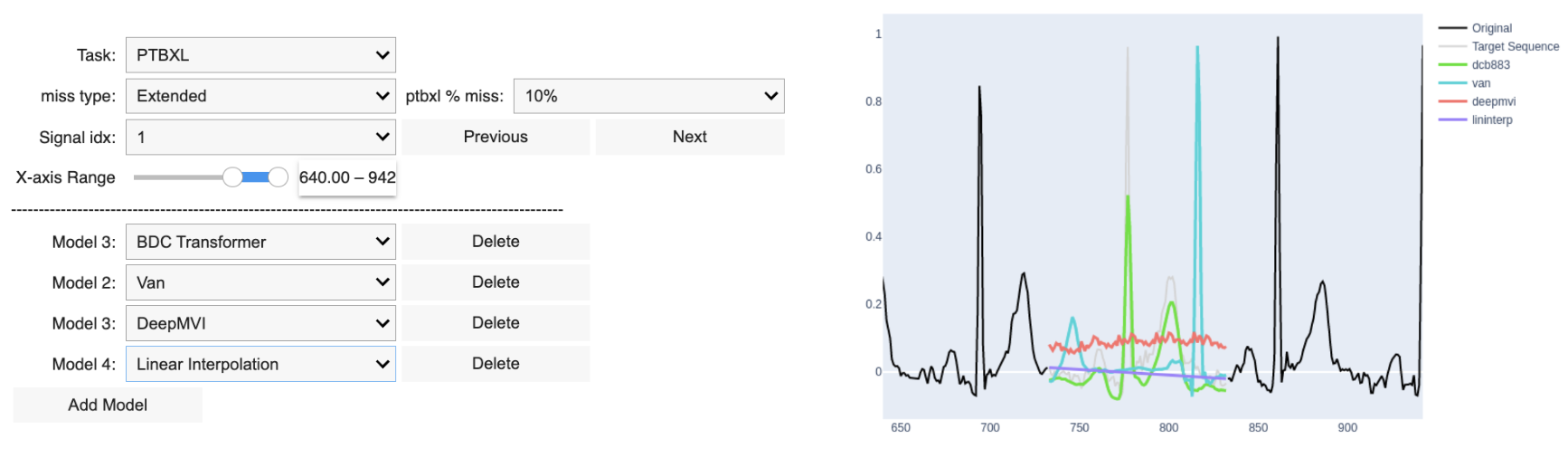}
    \centering
    \vspace{-.8cm}
    \caption{Visualization demo demonstrating the imputation results of BDC trasformer, Vanilla Transformer, DeepMVI, and Linear Interpolation, compared to the ground truth. The left shows the inputs for the interactive visualization and on the right is the plot.}
\label{figure:visual}
\end{figure}

\vspace{0.75em}
\noindent\textbf{Visualization}\indent 
The \texttt{Visualization} module accepts a dataset, missingness type, and any number of model output imputations as inputs. The output displays the original imputation, generated imputations, and actual signal value in a missing region for a given data sample. This module is offered as an interactive notebook, with buttons and dropdown menus to easily view different experiments, magnify areas of the plot, and make comparisons between imputation methods. A demo is provided with the results of fully trained models on a small subset of data samples, and can be seen in Fig. \ref{figure:visual}.

\section{Usage Examples}
PyPulse can be run by (i) optionally uploading a custom dataset, (ii) modifying or selecting the desired experiment configuration, and (iii) running the algorithm. When calling the runner, users can switch between various configurations. The following illustrates running the workflow for various experiments according to the user's needs:

\begin{lstlisting}[language=Python]
# evaluate SOTA method on custom dataset
$ python3 run.py -d customdatasetname
# re-train SOTA method on the MIMIC-III PPG dataset.
$ python3 run.py -c BDCTransformer/bdc_mimic_ppg_train.yaml
# train novel method on MIMIC-III PPG dataset. 
$ python3 run.py -c CustomModel/custom_train.yaml -d mimiciii_ppg
# evaluate novel method on custom dataset
$ python3 run.py -c CustomModel/custom_train.yaml -train False -d customdataname
\end{lstlisting}

Submodules of PyPulse can be run independently. For example, the visualize, evaluation, and missingness functions within run.py can be run as static functions. Below is an example of using the utils.visualize function independently to plot and save imputation results.

\begin{lstlisting}[language=Python]
utils.visualize(task='PTBXL', ptbxl1='Extended', ptbxl2='10%', 
        models=['BDCTransformer', 'FFT', 'NAOMI', 
        'VanillaTransformer'], sample_index=0, x_range=5000, 
        save_path='ptbxl_miss_extended10.png')
\end{lstlisting}

\section{Conclusion}
PyPulse equips researchers with a comprehensive set of tools to experiment with and refine novel imputation methods. In response to a need for imputation software, PyPulse aims to help health and machine learning researchers quickly compare imputation algorithms on a custom dataset or adapt the modular codebase for their needs. 

\bibliography{references}

\end{document}